\documentclass[10pt,twocolumn,letterpaper]{article}

\usepackage{cvpr}              % To produce the CAMERA-READY version
\usepackage{graphicx}
\usepackage{amsmath}
\usepackage{amssymb}
\usepackage{booktabs}
\usepackage{graphics}
\usepackage{multirow}
\usepackage{balance}
\usepackage[accsupp]{axessibility}

\usepackage[pagebackref,breaklinks,colorlinks]{hyperref}

\usepackage[capitalize]{cleveref}
\crefname{section}{Sec.}{Secs.}
\Crefname{section}{Section}{Sections}
\Crefname{table}{Table}{Tables}
\crefname{table}{Tab.}{Tabs.}

\def\cvprPaperID{19} % *** Enter the CVPR Paper ID here
\def\confName{CVPR}
\def\confYear{2023}

\begin{document}

\title{On the Real-Time Semantic Segmentation of Aphid Clusters in the Wild}
%\and
\author{Raiyan Rahman\textsuperscript{\rm 1}, 
    Christopher Indris\textsuperscript{\rm 1}
    Tianxiao Zhang\textsuperscript{\rm 2}\\
    Kaidong Li\textsuperscript{\rm 2},
    Brian McCornack\textsuperscript{\rm 3},
    Daniel Flippo\textsuperscript{\rm 4},
    Ajay Sharda\textsuperscript{\rm 4},
    Guanghui Wang\\[6pt]\textsuperscript{\rm 1}%\thanks{Corresponding author: Guanghui Wang (\email{wangcs@torontomu.ca})}\\[6pt]  
    \textsuperscript{\rm 1}Department of Computer Science, Toronto Metropolitan University, Toronto, Canada\\ 
    \textsuperscript{\rm 2}Department of Electrical Engineering and Computer Science, University of Kansas, Lawrence, USA\\
    \textsuperscript{\rm 3}Department of Entomology, Kansas State University, Manhattan, USA\\
    \textsuperscript{\rm 4}Department of Biological and Agricultural Engineering, Kansas State University, Manhattan, USA\\
}

\maketitle

%%%%%%%%% ABSTRACT
\begin{abstract}
   %Aphid infestations are known to cause massive damage to wheat and sorghum fields and are often vectors for plant viruses, resulting in major yield losses in the agricultural sector. To combat this, farmers employ the use of chemical pesticides which are inefficiently applied in large areas evenly across the fields, with much of it being wasted on areas that contain no pests while not applying sufficient amounts to areas with more severe infestations. Although automated pest detection and segmentation has been studied as a possible solution, to be able to employ this form of system in fields would require the ability to detect these infestations in real-time via autonomous systems. In this paper, we demonstrate the viability of real-time semantic segmentation of aphid clusters in the wild. We generate image patches across multiple scales and compare the accuracy and segmentation speeds of four state-of-the-art real-time semantic segmentation models on this aphid cluster dataset, as well as benchmark them against nonreal-time models. Our results show the effectiveness of a real-time solution which will aid in reducing inefficient pesticide use while helping to increase crop yields, laying the stepping stones towards an autonomous pest detection system.
   Aphid infestations can cause extensive damage to wheat and sorghum fields and spread plant viruses, resulting in significant yield losses in agriculture. To address this issue, farmers often rely on chemical pesticides, which are inefficiently applied over large areas of fields. As a result, a considerable amount of pesticide is wasted on areas without pests, while inadequate amounts are applied to areas with severe infestations. The paper focuses on the urgent need for an intelligent autonomous system that can locate and spray infestations within complex crop canopies, reducing pesticide use and environmental impact. We have collected and labeled a large aphid image dataset in the field, and propose the use of real-time semantic segmentation models to segment clusters of aphids. A multiscale dataset is generated to allow for learning the clusters at different scales. We compare the segmentation speeds and accuracy of four state-of-the-art real-time semantic segmentation models on the aphid cluster dataset, benchmarking them against nonreal-time models. The study results show the effectiveness of a real-time solution, which can reduce inefficient pesticide use and increase crop yields, paving the way towards an autonomous pest detection system.
\end{abstract}

%%%%%%%%% BODY TEXT
\section{Introduction}

The challenge of meeting the growing global demand for food under various environmental conditions has become increasingly significant. It is estimated that 37\% of crops worldwide are lost to pest damage, with 13\% attributed to insects. This loss includes staple crops such as rice, wheat, and maize, which directly impacts global food security. Additionally, commodity crops like banana and coffee have a significant impact on national economies, as shown in \cite{finegold2019global}. The increasing demand for food has led to the widespread use of chemical pesticides, which have become a critical component of crop production for maximizing yields. As a result, the global pesticide market was valued at approximately \$53.7 billion in 2015, and the economic value of global pesticide imports has been increasing by 7\% annually since 2000 \cite{shattuck2021generic}. Farmers typically apply uniform and continuous pesticide treatment across entire fields once a pest infestation reaches a specific treatment threshold. However, this approach is excessive as pest incidence and severity are only present in a fraction of the field. This highlights the urgent need for an intelligent autonomous system that can accurately locate and spray infestations within complex crop canopies, minimizing pesticide use and reducing environmental impact.

\begin{figure}
  \centering
  \includegraphics[width=0.9\columnwidth]{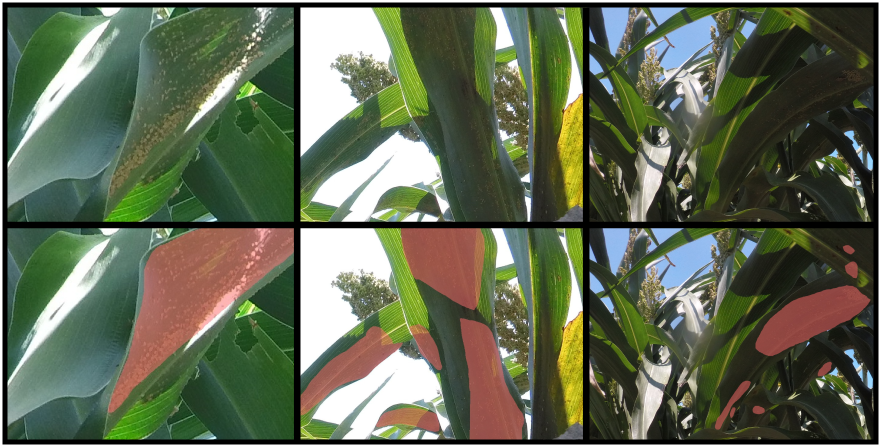}
  \caption{Examples of dataset images and their corresponding ground truth masks. The first, second, and third columns represent image patches at Scale-1, Scale-2, and Scale-3, respectively. The top row shows the aphid clusters while the bottom row shows their corresponding ground truth masks overlaid on top of them.}
  \label{fig:examples}
\end{figure}

The topic of automatic pest detection has been extensively studied, with significant challenges arising due to the small size of pests and their ability to blend into their natural settings. Previous studies have focused on counting individual pests \cite{teng2022td, chen2018automatic, barbedo2014using}, while not putting as much focus on the pressing issue of identifying infestations. In recent years, through the development of small autonomous vehicles such as drones, researchers have been able to study the viability of autonomous pest-scouting vehicles with camera feeds for the eventual goal of exploring crop canopies and identifying pest infestations. While work has been done on the detection and segmentation of aphids \cite{liu2016detection, chen2018automatic, teng2022td, zhang2023new}, none have focused on this real-time aspect which is crucial for an autonomous system. With the use of real-time semantic segmentation models, we explore the feasibility of accurate detection of pests in real-time as these vehicles with their feeds move throughout the fields.

In this paper, we focus on segmenting clusters of aphids in real-time using semantic segmentation models. We use high-resolution images captured from a sorghum field to create a multiscale dataset that allows for learning the clusters at different scales. We train our real-time models and compare their performance against state-of-the-art nonreal-time models, evaluating them across a wide range of metrics and segmentation speeds. To verify the effectiveness of different scales, we also train our models on each individual scale and compare their performance to the multiscale version of the dataset. Our goal is to identify the best model for automated pest infestation segmentation by exploring the relationship between accuracy and speed. Based on our results, Fast-SCNN and Small HRNet are two promising models that achieve high accuracy while also providing real-time segmentation speeds. The generated dataset can be accessed at: { https://doi.org/10.7910/DVN/N3YJXG}.

%-------------------------------------------------------------------------
\section{Related Works}

Computer vision techniques have been extensively used in the research of automated pest detection. Barbedo \cite{barbedo2014using} demonstrated an automatic method of detecting and counting pests on leaves with simple backgrounds. It was suggested that designing more effective features would further improve segmentation performance. A more robust aphid identification model was developed by \cite{liu2016detection} using histogram-oriented gradient features and a support vector machine. They utilized a maximally stable extremal region descriptor to allow the model to focus on pest regions rather than backgrounds. However, these methods required carefully designed features to describe and detect the targets. Thus, there was a movement towards designing models that could learn effective features from the data itself.

Convolutional neural networks (CNNs) are one of the most effective feature extraction and classification methods, achieving numerous state-of-the-art results in various image analysis tasks, such as classification \cite{ chen2022improving, ma2022semantic}, object detection \cite{li2021colonoscopy}, segmentation \cite{he2021sosd, patel22fuzzynet}, object counting \cite{ sajid2022towards}, and tracking \cite{ zhang2020efficient}. However, the detection of small insects can be especially challenging and even the state-of-the-art models have difficulty detecting individual aphids to satisfactory results. A modified U-Net architecture was proposed in \cite{chen2018automatic} to segment and count aphid nymphs on leaves. Although the model achieved high precision and recall and was generic to different species of pests, it was trained on ideal images with black backgrounds rather than the natural field settings in which such a detection system would be necessary. In these real-world settings, most aphids are clustered together on leaves as well as partially covered by adjacent foliage, making their individual detection amidst the dense clusters very difficult.

TD-Det \cite{teng2022td} was proposed for aphid detection with two core designs, a transformer-feature pyramid network (T-FPN) and a multi-resolution training method (MTM). Most of these existing methods attempt to identify individual aphids for counting and usually take place in ideal lighting conditions. The study by Zhang \etal \cite{zhang2023new} addressed the need for a dataset of pests within their natural habitat, creating a new aphid cluster detection dataset where images affected by aphids were manually selected and annotated. Instead of individual aphids, they were annotated in clusters to assess the overall infestation levels across the field. Bounding boxes were created based on these aphid clusters and the performance of four object detection models was benchmarked on square image patches of 400 pixels. However, the detection bounding box is not an ideal measure for infestation levels. As the importance of real-time segmentation systems is becoming more prevalent, the feasibility of such when it comes to identifying pests is ripe for study.

\begin{figure}
  \centering
  \includegraphics[width=1.0\columnwidth]{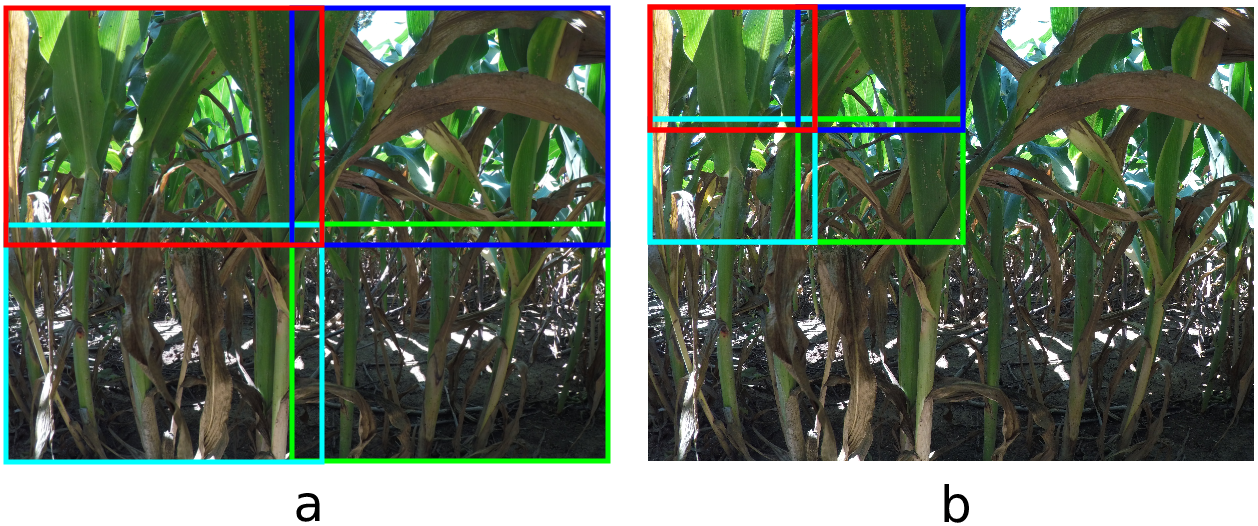}
  \caption{Example of two of the different scales that were used in training. Image (a) shows the original high-resolution image of resolution 3647$\times$2736 with patches at the 0.525W$\times$0.525H scale, where W and H refer to the width and height of the original image. At this scale, the original image will yield 4 patches. Image (b) shows how the patches were generated at 0.263W$\times$0.263H relative to the original image. This scale will yield 16 patches from the original image. In each case, adjacent patches were taken with an overlap of 10\%.}
  \label{fig:scales}
\end{figure}

\begin{figure*}[t]
\centering
\includegraphics[width=0.9\textwidth]{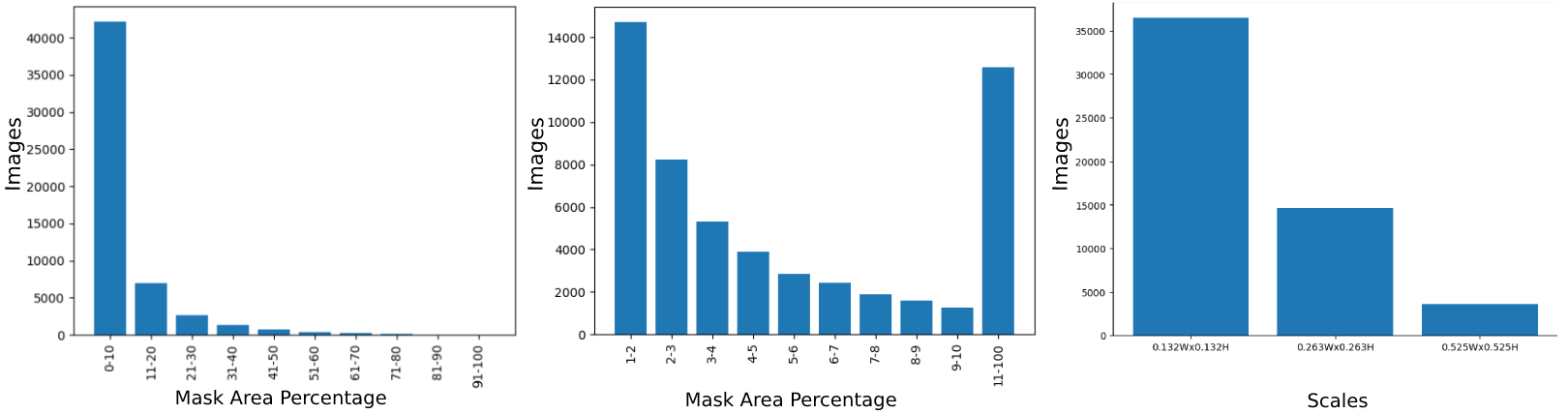}
\caption{Histograms showing the mask area percentage across the images as well as the number of images per scale. The chart on the left shows the percentage of aphid cluster masks using intervals of 10\%. As most take place in the interval of 0\% to 10\%, the chart in the center further breaks that interval down for better analysis. On the right, we can see the comparison between the number of images at each scale.}
\label{fig:mask-stats}
\end{figure*}

%-------------------------------------------------------------------------
\section{Dataset}

Using a three-camera GoPro rig at different heights, a variety of images were captured from a sorghum field over two seasons to gather images of aphid clusters. Most images did not contain aphids as they are localized in specific regions of the field, so images without aphids were filtered out, leaving 5,447 images. These were manually labeled by trained researchers to create segmentation masks, resulting in 59,767 labeled aphid cluster images. To ensure sufficient representation of aphid clusters, images with less than 1\% of pixels containing aphid clusters were filtered out, leaving a total of 54,742 images in the dataset.

In fields that are often plagued by aphid infestations, these pests tend to group closely together rather than being distributed individually. Therefore, to accurately assess the extent of infestation, it is more useful and efficient to monitor for clusters of aphids. However, it is necessary to set a proper threshold to ensure that a sufficient number of aphids are present to constitute a cluster. The study \cite{zhang2023new} defined any group of six or more aphids located close to one another as an aphid cluster. This approach ensures that if some aphids were too sparsely distributed, they would not be considered a significant economic threat. The aphid clusters in our dataset were also considered to be those with six or more aphids to meet the same threshold.

The original collected images consisted of very high-resolution images where most masks made up only 0.015\% of the images. The study \cite{zhang2023new} demonstrated the feasibility of utilizing square patches of 400 pixels with 50\% overlap between adjacent patches for the purpose of cluster detection. While this may aid models to better learn to detect the clusters, for the purpose of real-time segmentation on high-resolution images the models must be more robust to detecting aphid clusters at different scales. Thus, we propose to generate patches at different scales to allow the model to be able to better generalize to aphids and images captured at different scales. Patches were generated at scales of 0.132W$\times$0.132H (Scale-1), 0.263W$\times$0.263H (Scale-2), and 0.525W$\times$0.525H (Scale-3), where W and H refer to the width and height of the original image, respectively. An example of what these scales look like is shown in \cref{fig:scales}. In each case, 10\% overlap was used for adjacent patches for the completeness of masks. Patch generation was performed after fold separation to prevent data leakage across them. All three scales were combined into the multiscale dataset and separated into 10 sets of images to perform 10-fold cross validation.

The final dataset contained 35,140, 13,311, and 6,291 images taken from cameras at the top, middle, and bottom heights respectively and 54,742 images in total. From these images, there were 36,478, 14,628, and 3,636 images from Scale-1, Scale-2, and Scale-3, respectively. The percentage of aphid clusters in each image was also analyzed and can be seen in \cref{fig:mask-stats}. The large majority of these clusters are very small with an average cluster percentage of 2.45\%. 

%-------------------------------------------------------------------------

\begin{figure*}[t]
\centering
\includegraphics[width=0.85\textwidth]{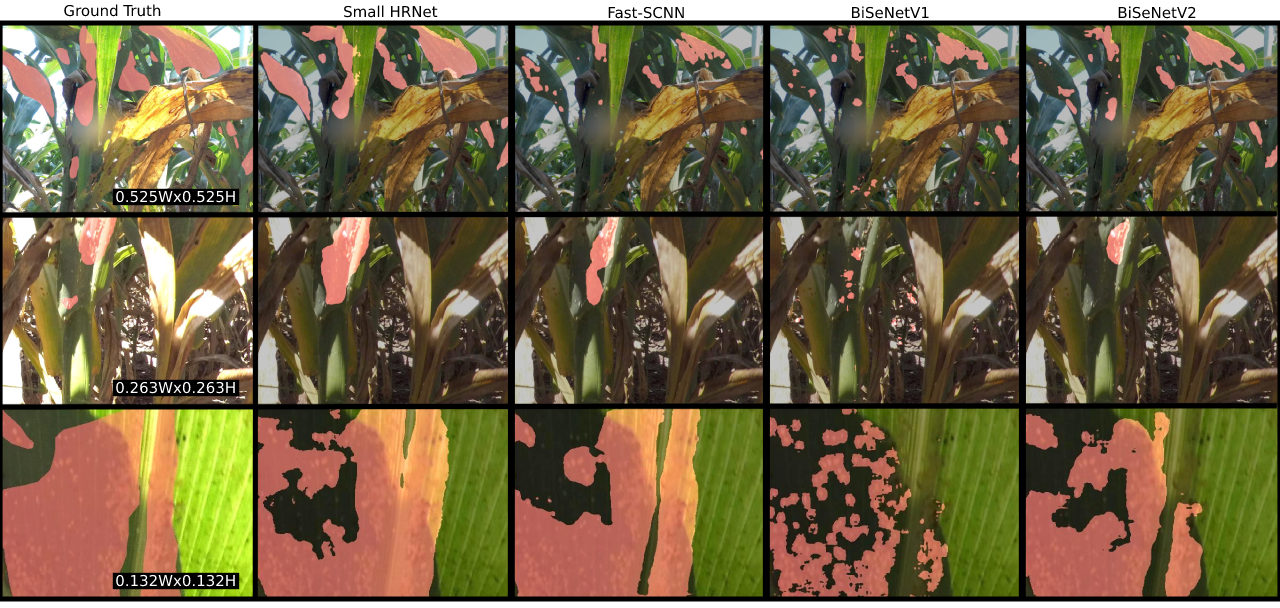}
\caption{Comparison between the ground truths and the segmentation results of the real-time segmentation models. The results of Small HRNet, Fast-SCNN, BiSeNetV1, and BiSeNetV2 are shown in the second, third, fourth, and fifth columns respectively. The three rows show segmentation results at Scale-3 (top), Scale-2 (middle), and Scale-1 (bottom).}
\label{fig:seg-results}
\end{figure*}

%-------------------------------------------------------------------------
\section{Real-Time Segmentation}

The development of real-time segmentation has been a significant factor in the increasing use of autonomous platforms, such as self-driving cars, robots, and drones. These platforms rely on the ability to make accurate predictions quickly, which is made possible through the low latency of real-time segmentation models. In 2016, the Efficient Neural Network (ENet) was introduced, which used efficient convolutions and a compact architecture to enable real-time semantic segmentation \cite{paszke2016enet}. This model was a significant breakthrough and inspired the development of other efficient and compact models for real-time computer vision applications.

Bilateral Segmentation Network (BiSeNet) \cite{yu2018bisenet} proposed to combine low-level features and high-level semantic information to find a good balance of speed and segmentation performance, achieving 68.4\% mIoU on high-resolution input from the Cityscapes test dataset at speeds of 105 FPS. A new version of this architecture, BiSeNetV2, was proposed in 2020 \cite{yu2021bisenet} and improved upon both the performance and the segmentation speed of its predecessor. This network managed to achieve 72.6\% mIoU on high-resolution input from the Cityscapes test set with faster speeds of 156 FPS. Fast-SCNN \cite{poudel2019fast} is another popular real-time segmentation model which achieves good segmentation speeds for high-resolution image data while also being suitable for embedded devices with low computational memory. The model achieves this by learning low-level features from multiple resolution branches simultaneously. On the Cityscapes dataset, Fast-SCNN is able to achieve an accuracy of 68\% mIoU at 123.5 FPS. In 2019, High-Resolution Network (HRNet) was introduced by \cite{wang2020deep} as a strong backbone for many computer vision problems where high-resolution representations are essential. It maintains high resolution throughout the whole process, resulting in semantically rich representations that are more spatially aware. A simpler version of this network, Small HRNet, managed to achieve real-time segmentation speeds at the slight cost of accuracy. It did so by using fewer layers and smaller width \cite{wang2020deep, yu2021lite}. The original HRNet achieved an accuracy of 80.9\% on the Cityscapes test set while its smaller variant achieved 73.86\% mIoU at the cost of faster speeds.

BiSeNetV1, BiSeNetV2, Fast-SCNN, and Small HRNet are all models that achieve real-time speeds across different datasets while also working with high-resolution images to obtain excellent performance. With the overall objective of an image analysis system for rapid pest identification and subsequent pest severity quantification at high spatial resolution, it is necessary for an approach that takes these into consideration. Thus, these four models were chosen for our problem of real-time aphid cluster segmentation using high-resolution images. %With a real-time segmentation solution, we ensured that segmentation of aphid clusters could take place at a sufficiently quick frame rate, allowing autonomous systems to utilize real-time information. However, from our search, the problem of real-time segmentation of pests in their natural setting has not been deeply studied. With this paper, we hope to garner interest in this problem space and demonstrate the possible efficacy of such a system.
By selecting a real-time segmentation solution, the ability to quickly segment aphid clusters in their natural setting is enhanced, allowing for efficient use of real-time information in autonomous systems. However, the real-time segmentation of pests in natural settings has not been extensively studied. This paper aims to draw attention to this challenge and demonstrate the potential effectiveness of such a system.

\begin{table*}
  \centering
  \begin{tabular}{@{}lclclclclc}
    \toprule
    Model & mIoU & mDice & mPrecision & mRecall & FPS \\
    \midrule
    HRNet-Small & \textbf{71.62 $\pm$ 0.47} & \textbf{81.15 $\pm$ 0.36} & \textbf{80.82 $\pm$ 1.20} & \textbf{81.64 $\pm$ 0.65} & 31.57\\
    Fast-SCNN & 71.25 $\pm$ 0.59 & 80.87 $\pm$ 0.50 & 80.46 $\pm$ 1.47 & 81.21 $\pm$ 0.67 & \textbf{91.66}\\
    BiSeNetV1 & 59.94 $\pm$ 0.54 & 69.22 $\pm$ 0.70 & 72.39 $\pm$ 1.55 & 67.12 $\pm$ 1.40 & 56.19\\
    BiSeNetV2 & 65.72 $\pm$ 0.53 & 75.58 $\pm$ 0.55 & 77.47 $\pm$ 1.34 & 74.06 $\pm$ 1.05 & 53.70\\
    \bottomrule
  \end{tabular}
  \caption{Real-time segmentation results from 10-fold cross-validation. \textbf{Bold} indicates top result.}
  \label{tab:rts-models}
\end{table*}

%-------------------------------------------------------------------------
\section{Experiments}

\textit{Pre-Processing:} The original dataset images were of very high resolution (3647$\times$2736), from which image patches were generated at three different scales, 0.132W$\times$0.132H, 0.263W$\times$0.263H, and 0.525W$\times$0.525H. These were then combined together and subsequently split into 10 sets for 10-fold cross validation.

As a pre-processing step, all images were resized to 1024$\times$768 prior to training. The images were also normalized according to the mean and standard deviation of the dataset. It was found that as most masks only consisted of a small percentage of the image, there is a large class imbalance between the background and aphid clusters. To account for this, the class weights were calculated from the pixels in the image masks as follows:

\begin{equation}
    W_{\textit{Aphid\;Cluster}} = \frac{\textit{Total\;Pixels}}{\textit{Aphid\;Cluster\;Pixels}}
\end{equation}

\begin{equation}
    W_{\textit{Background}} = \frac{\textit{Total\;Pixels}}{\textit{Background\;Pixels}}
\end{equation}

\textit{Training Set Up:} The models were all implemented with PyTorch in Python using the MMSegmentation framework \cite{mmseg2020}. They were trained on a Linux server with 4 NVIDIA Tesla P100 GPUs with 48G memory in total.

\textit{Training Pipeline:} Stochastic Gradient Descent (SGD) was used with a learning rate of 0.001, a momentum of 0.9, and a weight decay of 0.0005. The models were trained with a batch size of 2 for 160,000 iterations. To calculate the loss during training, we utilized the Cross-Entropy Loss and Dice loss with weights of 1 and 2 respectively.

\textit{Model Evaluation:} To evaluate our models, we used the intersection over union (IoU) metric, one of the foundational evaluation metrics in detection and segmentation, as well as the Dice score, a very popular evaluation metric for segmentation. These metrics are defined as follows:

\begin{equation}
    IoU = \frac{\textit{Area\;of\;Overlap}}{\textit{Area\;of\;Union}}
\end{equation}

\begin{equation}
    Dice = \frac{\textit{2} \times \textit{Area of Overlap}}{\textit{Total Area}}
\end{equation}

We also calculate the Precision and Recall of the models, and they are defined as follows:

\begin{equation}
    Precision = \frac{TP}{TP + FP}
\end{equation}

\begin{equation}
    Recall = \frac{TP}{TP + FN}
\end{equation}

To compare the speed with which the models are able to segment images to detect aphid clusters, we will be evaluating the frames per second (FPS) of each model.

\begin{table}
  \centering
  \resizebox{\columnwidth}{!}{%
  \begin{tabular}{@{}lclclclclc}
    \toprule
    Model & mIoU & mDice & mPrecision & mRecall & FPS \\
    \midrule
    PSPNet & \textbf{75.34} & \textbf{84.33} & 82.55 & \textbf{86.37} & 6.46\\
    DeepLabV3 & 75.05 & 84.06 & \textbf{83.21} & 83.92 & 3.86\\
    HRNet & 73.82 & 83.07 & 82.21 & 83.98 & \textbf{17.27}\\
    FCN & 72.43 & 81.88 & 81.43 & 82.35 & 6.43\\
    \bottomrule
  \end{tabular}%
  }
  \caption{Nonreal-time segmentation results. \textbf{Bold} indicates top results.}
  \label{tab:non-rts-models}
\end{table}

%-------------------------------------------------------------------------
\begin{figure*}
  \centering
  \includegraphics[width=0.8\textwidth]{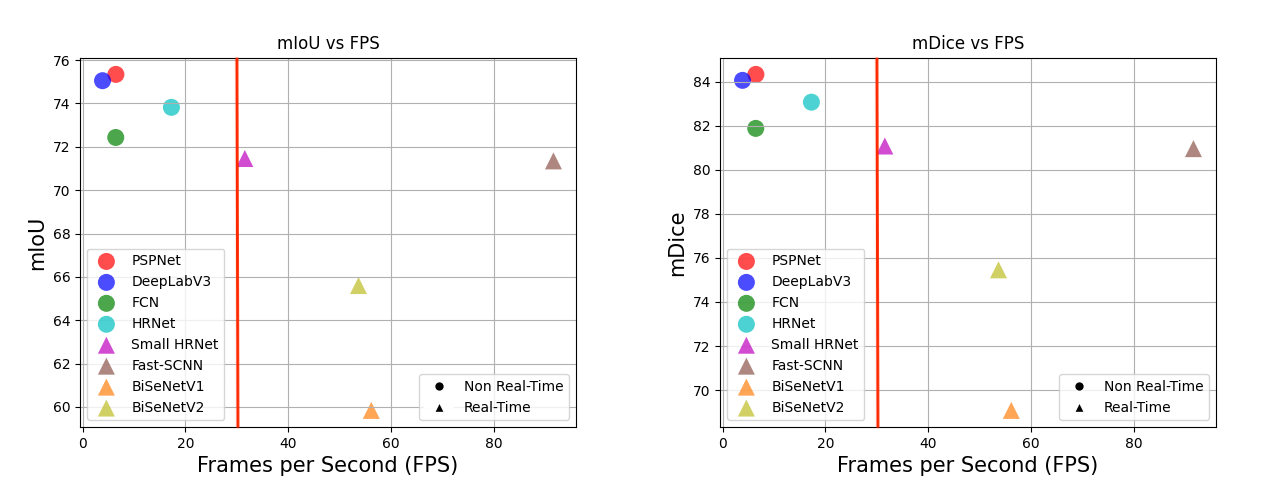}
  \caption{Scatter plots showing the accuracy vs the segmentation speed in frames per second (FPS). The plot on the left shows the accuracy as mean intersection over union while the right shows it as mean Dice score. The nonreal-time models are marked with circles while the real-time segmentation models are marked with triangles. The red line in each plot indicates 30 FPS which is considered the real-time speed threshold.}
  \label{fig:speed-vs-accuracy}
\end{figure*}

\begin{table*}
  \centering
  \resizebox{2.1\columnwidth}{!}{%
  \begin{tabular}{@{}l cccc cccc cccc}
    \toprule
    &
      \multicolumn{4}{c}{0.132H$\times$0.132W} &
      \multicolumn{4}{c}{0.263H$\times$0.263W} &
      \multicolumn{4}{c}{0.525H$\times$0.525W}\\
      \cmidrule(lr){2-5} \cmidrule(lr){6-9} \cmidrule(lr){10-13}
    Model & mIoU & mDice & mIoU(M) & mDice(M) & mIoU & mDice & mIoU(M) & mDice(M) & mIoU & mDice & mIoU(M) & mDice(M)\\
    \midrule
    HRNet-Small & 69.91 & 79.63 & 65.11 & 74.84 & 69.73 & 78.95 & 65.14 & 74.84 & 68.33 & 77.21 & 59.41 & 68.1 \\
    Fast-SCNN & 69.88 & 79.6 & 65.58 & 75.29 & 70.07 & 79.25 & 67.16 & 76.93 & 68.82 & 77.73 & 65.52 & 75.38 \\
    BiSeNetV1 & 59.77 & 68.97 & 57.60 & 66.07 & 64.64 & 73.64 & 60.39 & 69.53 & 65.02 & 73.57 & 55.75 & 63.29 \\
    BiSeNetV2 & 65.51 & 75.32 & 60.49 & 69.68 & 66.69 & 75.79 & 61.23 & 70.39 & 66.09 & 74.81 & 57.78 & 66.02 \\
    \bottomrule
  \end{tabular}%
  }
  \caption{Results of the real-time segmentation models across the individual scales. For each scale, mIoU and mDice give the accuracy when tested only on images at their respective scale while mIoU(M) and mDice(M) denote the accuracy when tested on the combined multiscale version of the dataset.}
  \label{tab:multiscale}
\end{table*}

\section{Results}

\subsection{Training Real-Time Models}

In our experiments, four state-of-the-art real-time semantic segmentation models were used to perform 10-fold cross-validation. We initially separated the original high-resolution images into separate groups and then generated patches from them at the three scales. We combined these generated patches into a single dataset, allowing the models to generalize to what aphid clusters look like at multiple scales. We then proceeded to train our models on this combined multiscale dataset. The performance of these models is shown in Table \ref{tab:rts-models}. The results show the mean intersection over union (mIoU), mean dice score (mDice), mean precision (mPrecision), mean recall (mRecall), and the speed in frames per second (FPS). As we can see, there is a usual trade-off between accuracy and speed. Small HRNet achieved the best accuracy while having a much slower speed compared to the other models at 31.57 FPS, albeit still in real-time. On the other hand, Fast-SCNN achieved excellent above real-time segmentation speeds of 91.66 FPS while maintaining a comparably good accuracy, only slightly lower than Small HRNet. The BiSeNetV1 model achieved the worst accuracy out of the chosen models but was able to achieve a respectable 56.19 FPS. BiSeNetV2 improved upon the performance of its predecessor while taking a slight blow to its speed. From these results, we have a clear distinction of the best-performing models with Small HRNet and Fast-SCNN. They are able to achieve the highest accuracy while also keeping above the 30 FPS real-time threshold.

\subsection{Comparison Against Nonreal-Time Models}

To provide adequate context regarding the segmentation accuracy obtained, four popular image segmentation models were also trained with the same hyperparameters. These models include the original version of HRNet, Pyramid Scene Parsing Network (PSPNet), DeepLabV3, and Fully Convolutional Network (FCN). These models are quite frequently used for many complicated semantic segmentation tasks and are chosen to allow us to analyze how accuracy and more complicated network structures would impact segmentation speed. These models were trained on the same generated dataset containing images at multiple scales. Their performance is shown in \Cref{tab:non-rts-models}. It is evident that the best performance was obtained by PSPNet. The performance of HRNet is also shown to be slightly better than its small counterpart, with just over half the speed of Small HRNet at 17.27 FPS. DeepLabV3 is able to achieve very similar performance to PSPNet while FCN, a relatively older network, achieves a lower accuracy compared to the other nonreal-time models but is able to still outperform all the real-time models. The segmentation speed of each model is also noted. While these models may achieve superior accuracy, they are unsuitable for our need for real-time segmentation, unlike the real-time models. Thus, through small sacrifices in accuracy, we are able to achieve real-time segmentation.

\subsection{Comparison of Different Multiple Scales}

In the study \etal \cite{zhang2023new}, square patches of 400 pixels were used to create the aphid clusters dataset for the training of object detection. This was due to the original images having a very small percentage of them containing aphid clusters. In this paper, we wanted to study the effectiveness of using multiple scales to have a generalizable representation of aphid clusters for semantic segmentation. We trained our real-time segmentation models on each of the individual generated scales to see how well they would perform not only on their own scales but on images of other scales as well through the combined dataset. The performance of these models is shown in \Cref{tab:multiscale}. For each scale, we show the accuracy for images at their own respective scale in terms of mIoU and mDice in the first two columns. The next two columns labeled mIoU(M) and mDice(M) show the accuracy on the combined dataset containing multiple scales. For Small HRNet and Fast-SCNN, we see a decrease in the overall accuracy across the scales as well as a significant decrease in accuracy when applied to the combined dataset. For the two BiSeNet models, we see a rise in accuracy at their respective scales showing that they are able to learn individual scales better. This is especially true for BiSeNetV1 which is also able to achieve a similar result at the 0.263 scale as it is able to achieve on the experiments using the combined dataset from \Cref{tab:rts-models}. However, for every other model, there is still a noticeable degradation in the accuracy for each scale when applied to the combined dataset. Thus, the use of the multiscale dataset has contributed to an increase in performance for almost all models.

\subsection{Segmentation Results}

\Cref{fig:seg-results} shows the segmentation results in comparison to the ground truths. The segmentation results of Small HRNet matched the ground truth mask most closely. It is able to detect most clusters that are sparsely located throughout the crop canopies, partially occluded, and also hidden by shadows. Small HRNet is also able to more accurately handle complicated boundaries. Fast-SCNN, on the other hand, creates tighter boundaries around the clusters and is able to avoid segmenting areas within and around the cluster that are only leaves without any aphids. In contrast to this, BiSeNetV1 is able to detect the general area of the aphid clusters but it also erroneously confuses other similarly textured parts of crops and flowers as clusters. BiSeNetV2 improves upon this and is able to create larger segmentation boundaries around the clusters and avoids segmenting other similarly textured parts of the crops. Both BiSeNetV1 and BiSeNetV2 are not as robust at handling different lighting conditions as Small HRNet and Fast-SCNN.

\begin{figure}
  \centering
  \includegraphics[width=0.98\linewidth]{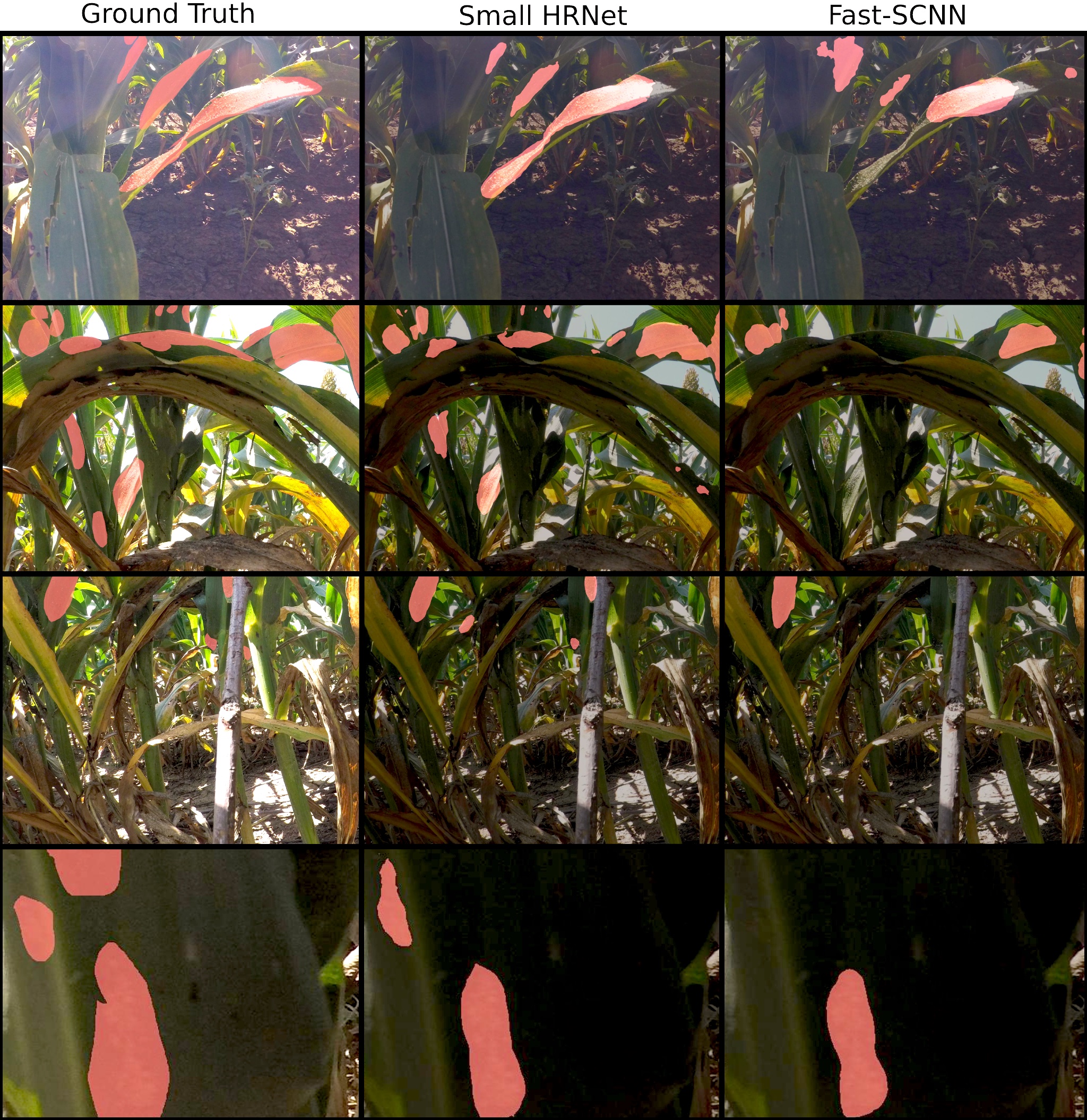}
  \caption{Further segmentation results from Small HRNet and Fast-SCNN alongside the ground truth images. From the different lighting conditions, blur, and cluster sizes we can make a better comparison between these two better-performing models.}
  \label{fig:comparison}
\end{figure}

%-------------------------------------------------------------------------
\section{Discussion}

Throughout the original collected images, aphid clusters usually only take up a very small percentage of each high-resolution image. It is important to ensure that models are robust to what these aphid clusters look like at different scales, at high resolutions with very small features, as well as at lower resolutions where larger features are able to be made out. Through the training of models with a variety of scales, we ensure they are able to generalize to changes in scale more effectively as seen through the increase in performance of the combined dataset compared to the individual scaled datasets.

From our results, apart from Fast-SCNN, there is a clear trade-off between accuracy and speed. This is especially evident in the case of HRNet and Small HRNet. By reducing the depth and the width and updating its stem to consist of two 3$\times$3 convolutions with a stride of 2 as stated in \cite{yu2021lite, holder2022efficient}, Small HRNet is able to improve upon the speed from 17.27 FPS to 31.57 FPS with only a small decrease in the accuracy. Fast-SCNN managed to achieve excellent speeds through the downsampling of images and the efficient sharing of computed low-level features. However, the simplicity of its architecture holds its performance back slightly behind Small HRNet.

Both Small HRNet and Fast-SCNN are able to perform very well with similar segmentation results. To demonstrate some key differences between them, \Cref{fig:comparison} shows some segmentation results of the two better-performing models where they had a big difference in their segmentation results. From the examples in this figure, we can see that smaller clusters are better detected by Small HRNet. Small HRNet is also able to better handle different lighting conditions such as darker regions underneath the canopies. However, Fast-SCNN is able to create tighter boundaries around the clusters compared to the much larger ones from Small HRNet. While this is better for smaller clusters or those that do not have many complicated shapes, it often misses surrounding smaller aphid subclusters that are present but separated from the main cluster body. Fast-SCNN is also able to better handle blurry images containing clusters.

Considering accuracy, both Small HRNet and Fast-SCNN are good candidates for the ideal choice. While the former has a slightly better performance, when considering the efficient nature of Fast-SCNN, its ability to find tighter boundaries of smaller clusters, and its very high framerate, it distinguishes itself as the ideal choice for the purpose of real-time aphid cluster segmentation.

%-------------------------------------------------------------------------
\section{Conclusion}

%The problem space of automated pest detection has seen a steady increase in interest over the past few years. High-resolution images were collected and rather than individual aphids, clusters of them were labelled to better determine the severity of infestations in the wild. In this paper, we generated a dataset for semantic segmentation using three different scales to aid in better generalization.  We proceeded to train four real-time segmentation models to assess how they would perform relative to state-of-the-art and popular semantic segmentation models while dealing with high-resolution input. The experiments demonstrate a general trade-off between the segmentation accuracy and speed due to complicated network structure. Through these results, we can ascertain that Fast-SCNN is the ideal choice due to its efficient nature and above real-time segmentation speed, with Small HRNet also being a good candidate with slightly better accuracy at the cost of speed. Automated pest control systems would be able to employ these real-time models to aid farmers employ pesticides in a more efficient way as well as limiting the overall amount of pesticide exposure of crops. With these promising results, we hope to inspire further research in the space of automated pest control and take us one step closer to a more efficient system overall.
Automated pest detection has become an increasingly popular research problem in recent years. In this study, we have collected high-resolution images and manually labeled clusters of aphids to better assess infestation severity in natural settings. We have created a dataset for semantic segmentation using three different scales to enhance generalization and trained four real-time segmentation models to evaluate their performance relative to state-of-the-art and popular semantic segmentation models on high-resolution input. Our experiments showed a trade-off between segmentation accuracy and speed due to the complexity of the network structure. Fast-SCNN emerged as the optimal choice, with real-time segmentation speed and efficient performance. Small HRNet also showed promise with slightly better accuracy, albeit at a cost of speed. Our findings suggest that automated pest control systems can use these real-time models to efficiently and accurately aid farmers in pesticide application, minimizing crop exposure to pesticides. These promising results motivate further research in automated pest control, bringing us closer to a more efficient and sustainable agricultural system.

% Update the cvpr.cls to do the following automatically.
% For this citation style, keep multiple citations in numerical (not
% chronological) order, so prefer \cite{Alpher03,Alpher02,Authors14} to
% \cite{Alpher02,Alpher03,Authors14}.
%------------------------------------------------------------------------
%%%%%%%%% REFERENCES
\balance
{\small
\bibliographystyle{ieee_fullname}
\bibliography{egbib}

\begin{thebibliography}{10}\itemsep=-1pt

\bibitem{barbedo2014using}
Jayme Garcia~Arnal Barbedo.
\newblock Using digital image processing for counting whiteflies on soybean
  leaves.
\newblock {\em Journal of Asia-Pacific Entomology}, 17(4):685--694, 2014.

\bibitem{chen2018automatic}
Jian Chen, Yangyang Fan, Tao Wang, Chu Zhang, Zhengjun Qiu, and Yong He.
\newblock Automatic segmentation and counting of aphid nymphs on leaves using
  convolutional neural networks.
\newblock {\em Agronomy}, 8(8):129, 2018.

\bibitem{chen2022improving}
Xiangyu Chen, Ying Qin, Wenju Xu, Andr{\'e}s~M Bur, Cuncong Zhong, and Guanghui
  Wang.
\newblock Improving vision transformers on small datasets by increasing input
  information density in frequency domain.
\newblock In {\em IEEE/CVF International Conference on Computer Vision
  Workshops (ICCVW)}, 2022.

\bibitem{mmseg2020}
MMSegmentation Contributors.
\newblock {MMSegmentation}: Openmmlab semantic segmentation toolbox and
  benchmark.
\newblock \url{https://github.com/open-mmlab/mmsegmentation}, 2020.

\bibitem{finegold2019global}
Cambria Finegold, Jeffrey Ried, Katherine Denby, and Sarah Gurr.
\newblock Global burden of crop loss.
\newblock {\em Gates Open Res}, 3, 2019.

\bibitem{he2021sosd}
Lei He, Jiwen Lu, Guanghui Wang, Shiyu Song, and Jie Zhou.
\newblock Sosd-net: Joint semantic object segmentation and depth estimation
  from monocular images.
\newblock {\em Neurocomputing}, 440:251--263, 2021.

\bibitem{holder2022efficient}
Christopher~J Holder and Muhammad Shafique.
\newblock On efficient real-time semantic segmentation: a survey.
\newblock {\em arXiv preprint arXiv:2206.08605}, 2022.

\bibitem{li2021colonoscopy}
Kaidong Li, Mohammad~I Fathan, Krushi Patel, Tianxiao Zhang, Cuncong Zhong,
  Ajay Bansal, Amit Rastogi, Jean~S Wang, and Guanghui Wang.
\newblock Colonoscopy polyp detection and classification: Dataset creation and
  comparative evaluations.
\newblock {\em Plos one}, 16(8):e0255809, 2021.

\bibitem{liu2016detection}
Tao Liu, Wen Chen, Wei Wu, Chengming Sun, Wenshan Guo, and Xinkai Zhu.
\newblock Detection of aphids in wheat fields using a computer vision
  technique.
\newblock {\em Biosystems Engineering}, 141:82--93, 2016.

\bibitem{ma2022semantic}
Wenchi Ma, Xuemin Tu, Bo Luo, and Guanghui Wang.
\newblock Semantic clustering based deduction learning for image recognition
  and classification.
\newblock {\em Pattern Recognition}, 124:108440, 2022.

\bibitem{paszke2016enet}
Adam Paszke, Abhishek Chaurasia, Sangpil Kim, and Eugenio Culurciello.
\newblock Enet: A deep neural network architecture for real-time semantic
  segmentation.
\newblock {\em arXiv preprint arXiv:1606.02147}, 2016.

\bibitem{patel22fuzzynet}
Krushi~Bharatbhai Patel, Fengjun Li, and Guanghui Wang.
\newblock Fuzzynet: A fuzzy attention module for polyp segmentation.
\newblock In {\em NeurIPS'22 Workshop on All Things Attention: Bridging
  Different Perspectives on Attention}.

\bibitem{poudel2019fast}
Rudra~PK Poudel, Stephan Liwicki, and Roberto Cipolla.
\newblock Fast-scnn: Fast semantic segmentation network.
\newblock {\em arXiv preprint arXiv:1902.04502}, 2019.

\bibitem{sajid2022towards}
Usman Sajid and Guanghui Wang.
\newblock Towards more effective prm-based crowd counting via a
  multi-resolution fusion and attention network.
\newblock {\em Neurocomputing}, 474:13--24, 2022.

\bibitem{shattuck2021generic}
Annie Shattuck.
\newblock Generic, growing, green?: The changing political economy of the
  global pesticide complex.
\newblock {\em The Journal of Peasant Studies}, 48(2):231--253, 2021.

\bibitem{teng2022td}
Yue Teng, Rujing Wang, Jianming Du, Ziliang Huang, Qiong Zhou, and Lin Jiao.
\newblock Td-det: A tiny size dense aphid detection network under in-field
  environment.
\newblock {\em Insects}, 13(6):501, 2022.

\bibitem{wang2020deep}
Jingdong Wang, Ke Sun, Tianheng Cheng, Borui Jiang, Chaorui Deng, Yang Zhao,
  Dong Liu, Yadong Mu, Mingkui Tan, Xinggang Wang, et~al.
\newblock Deep high-resolution representation learning for visual recognition.
\newblock {\em IEEE transactions on pattern analysis and machine intelligence},
  43(10):3349--3364, 2020.

\bibitem{yu2021bisenet}
Changqian Yu, Changxin Gao, Jingbo Wang, Gang Yu, Chunhua Shen, and Nong Sang.
\newblock Bisenet v2: Bilateral network with guided aggregation for real-time
  semantic segmentation.
\newblock {\em International Journal of Computer Vision}, 129:3051--3068, 2021.

\bibitem{yu2018bisenet}
Changqian Yu, Jingbo Wang, Chao Peng, Changxin Gao, Gang Yu, and Nong Sang.
\newblock Bisenet: Bilateral segmentation network for real-time semantic
  segmentation.
\newblock In {\em Proceedings of the European conference on computer vision
  (ECCV)}, pages 325--341, 2018.

\bibitem{yu2021lite}
Changqian Yu, Bin Xiao, Changxin Gao, Lu Yuan, Lei Zhang, Nong Sang, and
  Jingdong Wang.
\newblock Lite-hrnet: A lightweight high-resolution network.
\newblock In {\em Proceedings of the IEEE/CVF conference on computer vision and
  pattern recognition}, pages 10440--10450, 2021.

\bibitem{zhang2023new}
Tianxiao Zhang, Kaidong Li, Xiangyu Chen, Cuncong Zhong, Bo Luo, Ivan~Grijalva
  Teran, Brian McCornack, Daniel Flippo, Ajay Sharda, and Guanghui Wang.
\newblock A new dataset and comparative study for aphid cluster detection.
\newblock In {\em 2nd AAAI Workshop on AI for Agriculture and Food Systems},
  2023.

\bibitem{zhang2020efficient}
Tianxiao Zhang, Xiaohan Zhang, Yiju Yang, Zongbo Wang, and Guanghui Wang.
\newblock Efficient golf ball detection and tracking based on convolutional
  neural networks and kalman filter.
\newblock {\em arXiv preprint arXiv:2012.09393}, 2020.

\end{thebibliography}
}

\end{document}